%% file: Neuroimage_HARDIReg_template.tex
\documentclass[12pt]{elsarticle}

\usepackage{times}
\usepackage{epsfig}
\usepackage{graphicx}
\usepackage{amsmath}
\usepackage{amssymb}
\usepackage{amsthm}
\usepackage{makeidx}  
\usepackage{url}
\usepackage{algorithm}
\usepackage{bm}
\usepackage{microtype}
\usepackage{graphicx}
\usepackage{url}
\usepackage{caption}
\usepackage{subfigure}
\usepackage[rightcaption]{sidecap}
\usepackage{pdflscape}
\usepackage[bookmarks=true]{hyperref}
\usepackage{lineno}
\usepackage{ulem}
\usepackage{afterpage}
\usepackage[usenames,dvipsnames]{color}

\input{HARDI_NIMG2011_macro}
\journal{NeuroImage}

\begin{document}
\begin{frontmatter}

\title{Bayesian Estimation of White Matter Atlas from High Angular Resolution Diffusion Imaging}

\author[l1]{Jia Du}
\author[l2]{Alvina Goh}
\author[l1,l3,l4]{Anqi Qiu\corref{cor1}}
\ead{bieqa@nus.edu.sg}

\cortext[cor1]{Correspondence to: Anqi Qiu, Department of Bioengineering,
National University of Singapore, 9 Engineering Drive 1, Block EA
03-12, Singapore 117576. Tel: +65 6516
7002. Fax: +65 6872 3069}
\address[l1]{Department of Bioengineering, National University of Singapore, Singapore}
\address[l2]{Department of Mathematics, National University of Singapore, Singapore}
\address[l3]{Singapore Institute for Clinical Sciences, Agency for Science, Technology and Research, Singapore}
\address[l4]{Clinical Imaging Research Center, National University of Singapore, Singapore}

\newpage
\begin{abstract}
We present a Bayesian probabilistic model to estimate the brain white matter atlas from high angular resolution diffusion imaging (HARDI) data. This model incorporates a shape prior of the white matter anatomy and the likelihood of individual observed HARDI datasets. We first assume that the atlas is generated from a known hyperatlas through a flow of diffeomorphisms and its shape prior can be constructed based on the framework of large deformation diffeomorphic metric mapping (LDDMM). LDDMM characterizes a nonlinear diffeomorphic shape space in a linear space of initial momentum uniquely determining diffeomorphic geodesic flows from the hyperatlas. Therefore, the shape prior of the HARDI atlas can be modeled using a centered Gaussian random field (GRF) model of the initial momentum. In order to construct the likelihood of observed HARDI datasets, it is necessary to study the diffeomorphic transformation of individual observations relative to the atlas and the probabilistic distribution of orientation distribution functions (ODFs). To this end, we construct the likelihood related to the transformation using the same construction as discussed for the shape prior of the atlas. The probabilistic distribution of ODFs is then constructed based on the ODF Riemannian manifold. We assume that the observed ODFs are generated by an exponential map of random tangent vectors at the deformed atlas ODF. Hence, the likelihood of the ODFs can be modeled using a GRF of their tangent vectors in the ODF Riemannian manifold. We solve for the maximum a posteriori using the Expectation-Maximization algorithm and derive the corresponding update equations. Finally, we illustrate the HARDI atlas constructed based on a Chinese aging cohort of $94$ adults and compare it with that generated by averaging the coefficients of spherical harmonics of the ODF across subjects. 
\end{abstract}

\end{frontmatter}

\newpage
\section{Introduction}
\label{sec:intro}

The white matter region of the human brain is composed of neuronal axons that provide insights on brain connections. Such information is very useful for identifying neuropathology of mental illnesses and understanding fundamental neuroscience questions on how the brain regions interact each other. Up to now, a comprehensive atlas that well characterizes the {\it in-vivo} white matter anatomy of the human brain and can be used in atlas-based neuroimaging research remains lacking.

In the last decade, diffusion weighted magnetic resonance imaging (DW-MRI) technique has exploited the property that water molecules move faster along neural axons than against them. By measuring water diffusion in the brain, the location and trajectories of axons can be visualized and the axonal pathways can be reconstructed using DW-MRI. Diffusion tensor imaging (DTI), where axonal orientations are modeled using a three-dimensional ellipsoid tensor, has become one of the most popular mathematical models to study the white matter axonal orientation. DTI has since become a fundamental tool that enables researchers to obtain a deep understanding of the human brain. 

There have been several different approaches to DTI atlas construction, either using scalar registration \citep{Goodlett:NeuroImage2009,Verhoeven:HBM2010}, multi-channel methods \citep{Park:NeuroImage2003} or by directly optimizing tensor similarity \citep{Zhang:MICCAI2007}. Since then, a large body of research has leveraged the DTI atlas information. Some of these provided anatomical validation \citep{Lawes:neuro2008} and anatomical labeling of fiber tracts 
\citep{mori10,KHua:neuro08}. The comprehensive work by \citet{mori10} provides a three-dimensional and two-dimensional in-vivo atlas of various white matter tracts in the human brain based on DTI and has become an essential resource for neuroimaging researchers. \citet{KHua:neuro08}
create a white matter parcellation atlas based on probabilistic maps of the major white matter tracts and show that there is an excellent correlation of fractional anisotropy and mean diffusivity between the automated and the individual tractography-based results. \citet{Lawes:neuro2008} show that it is possible to establish a close correspondence of the fiber tracts generated from the DTI atlas with the tracts isolated with classical dissection of post-mortem brain tissue. A DTI atlas containing the complete diffusion tensor information is constructed by \citet{Verhoeven:HBM2010}. Using robust fiber tracking methods on this DTI atlas, Verhoeven et al. reconstruct a large number of white matter tracts and show that their framework yields highly reproducible and reliable fiber tracts. \citet{ThiebautdeSchotten:Neuro2011} produce a white matter atlas that describes the in-vivo variability of the
major association, commissural, and projection connections and study the
inter-subject variability between left and right hemispheres in relation to
gender based on this atlas.  DTI atlases have been directly used to study white matter fiber tracts \citep{ODonnell:NeuroImage2009,Yushkevich:NeuroImage2008}.


It has been demonstrated that DTI is valuable for studying brain white matter development in children and detecting abnormalities in patients with neuropsychiatric disorders and neurodegenerative diseases \citep[e.g.][]{Huang:Neuroimage,NBM:NBM789,Qiu:PLoS_AD}. However, a major shortcoming of DTI is that it can only reveal one dominant axonal orientation at each location while between one and two thirds of the human brain white matter are thought to contain multiple axonal bundles crossing each other \citep{Behrens:NeuroImage07}. In contrast, HARDI \citep{Tuch:MRM02b} addresses this well-known limitation of DTI by measuring water diffusion along uniformly distributed directions on the sphere. It can thus characterize more complex axonal geometries. HARDI measurements are used to reconstruct an orientation distribution function (ODF), a probability density function (PDF) defined on the sphere, to describe the axonal distribution. Unlike the tensor model used in DTI, the ODF has no restriction on the number of axons present in a specific anatomical location and thus can well characterize the true underlying white matter architecture. By quantitatively comparing axonal orientations retrieved from the ODFs against histological measurements, \citet{leergaard_plosone_2010} show that accurate estimates of axon bundles can be obtained from HARDI data, therefore further validating its usage in brain studies.

Over the last decade, atlas generation techniques based on intensity images have matured significantly and they include those based only on affine or non-linear registration methods \citep{joshi-davis-jomier-gerig-2004,avants-gee-2004} and probabilistic models coupled to the Expectation-Maximization (EM) algorithm to estimate both a shape prior of the atlas and an intensity image likelihood function \citep{steph_template_2007,jun_nimg_2008,Qiu:TIP10}. In contrast, the white matter atlas generation based on HARDI is still very much in its infancy. \citet{BouixMICCAI2010} employ an image registration approach that first seeks the transformation between fractional anisotropic (FA) images and then resample the HARDI signals of each subject into a common coordinate according to this transformation. The HARDI atlas is then generated by averaging the coefficients of spherical harmonics of the ODF across subjects. \citet{Yeh201191} first construct the spatial normalization of the diffusion information using a q-space diffeomorphic reconstruction method, reconstruct the spin distribution function (SDF) in the ICBM-$152$ space from the diffusion MR signals, and the white matter atlas is then computed by averaging the SDF over individual subjects. \citet{Bloy2011} perform alignment of ODF fields by using a multi-channel diffeomorphic demons registration algorithm on rotationally invariant feature maps and white matter parcellation is done via a spatially coherent normalized cuts algorithms. 

To the best of our knowledge, there is no probabilistic framework for generating the HARDI atlas that incorporates both a shape prior of the white matter anatomy and a probabilistic model of the ODFs. In this paper, we extend the previous Bayesian model for the intensity image atlas generation proposed in \citep{jun_nimg_2008,Qiu:TIP10} to that for HARDI. Briefly, we derive a Bayesian model with a shape priori of the HARDI atlas in terms of diffeomorphic transformations and a likelihood function of the ODFs in terms of their tangent vectors on an ODF Riemannian manifold. As we will see later, the extension of the Bayesian model from intensity images to HARDI is non-trivial. Our main contributions of this work are to construct the likelihood function of the ODFs based on their Riemannian structure and derive the Expectation-Maximization algorithm and the update equations for solving the Bayesian HARDI atlas estimation. In the following methodological sections, we first introduce the general framework of this Bayesian HARDI atlas estimation in \S \ref{subsec:generalframework} and construct the shape prior of the atlas and the distribution of random diffeomorphisms given the estimated atlas. In \S \ref{subsec:likelihood}, we construct the conditional likelihood of ODFs based on their Riemannian manifold. In \S \ref{subsec:em}, we derive the Expectation-Maximization algorithm to obtain the maximum a posteriori solution and \S \ref{secsub:update} gives the proof of the EM update equations. We employ this atlas generation approach on $94$ HARDI datasets acquired in a Chinese aging study and \S \ref{sec:experiments} illustrates the estimated HARDI atlas. Our findings show that the atlas estimated using our algorithm preserves anatomical details of the white matter. As age increases, the corpus callosum thinning was observed, which is consistent with existing literature, \citep[e.g.][]{Hinkle:preprint2012,Fletcher:MFCA2011,Sullivan:CC2002}. Additionally, we demonstrate age effects on crossing fiber regions. Last but not least, we also compare our method to an existing HARDI atlas generation method  by averaging the coefficients of spherical harmonics of the ODF across subjects \citep{Bloy2011}.

\section{Methods}
\subsection{General Framework of Bayesian HARDI Atlas Estimation}
\label{subsec:generalframework}

In this section, we introduce the general framework of the Bayesian HARDI atlas estimation. Given $n$ observed ODF datasets $J^{(i)}$ for $i=1,\dots, n$, we assume that each of them can be estimated through an unknown atlas $I_{\atlas}$ and a diffeomorphic transformation $\phi^{(i)}$ such that 
\begin{align}
J^{(i)} \approx I^{(i)}= \phi^{(i)} \cdot I_{\atlas}.
\end{align}
The total variation of $J^{(i)}$ relative to $I^{(i)}$ is then denoted by $\sigma^2$. The goal here is to estimate the unknown atlas $I_{\atlas}$ and the variation $\sigma^2$. To solve for the unknown atlas $I_{\atlas}$, we first introduce an ancillary ``hyperatlas'' $I_0$, and assume that our atlas is generated from it via a diffeomorphic transformation of $\phi$ such that $I_{\atlas}=\phi \cdot I_0$. We use the Bayesian strategy to estimate $\phi$ and $\sigma^2$ from the set of observations $J^{(i)}, i=1,\dots, n$ by computing the maximum a posteriori (MAP) of $f_{\sigma}(\phi |J^{(1)}, J^{(2)},\dots, J^{(n)}, I_0)$. This can be achieved using the Expectation-Maximization algorithm by first computing the log-likelihood of the complete data ($\phi, \phi^{(i)}, J^{(i)},i=1,2,\dots,n$) when $\phi^{(1)}, \cdots, \phi^{(n)}$ are introduced as hidden variables. We denote this likelihood as $f_{\sigma}( \phi, \phi^{(1)}, \dots, \phi^{(n)}, J^{(1)},\dots J^{(n)}| I_0)$. We consider that the paired information of individual observations, $(J^{(i)}, \phi^{(i)})$ for  $i=1,\dots, n$, as independent and identically distributed. As a result, this log-likelihood can be written as 
\begin{align}
\label{eqn:loglikelihood}
&  \log f_{\sigma}( \phi, \phi^{(1)}, \dots, \phi^{(n)}, J^{(1)},\dots J^{(n)} | I_0) \\ \nonumber
=\; & \log f(\phi | I_0) + \sum_{i=1}^n \Big\{ \log f(\phi^{(i)} | \phi, I_0) + \log f_{\sigma}(J^{(i)}|\phi^{(i)}, \phi, I_0) 
\Big\}  \ ,
\end{align}
where $f(\phi | I_0)$ is the shape prior (probability distribution) of the atlas given the hyperatlas, $I_0$. $f(\phi^{(i)} | \phi, I_0)$ is the distribution of random diffeomorphisms given the estimated atlas ($\phi \cdot I_0$). $f_{\sigma}(J^{(i)}|\phi^{(i)}, \phi, I_0)$ is the conditional likelihood of the ODF data given its corresponding hidden variable $\phi^{(i)}$ and the estimated atlas ($\phi \cdot I_0$). In the remainder of this section, we first adopt $f(\phi | I_0)$ and $f(\phi^{(i)} | \phi, I_0)$ introduced in \citep{jun_nimg_2008,Qiu:TIP10} and then describe how to calculate $f_{\sigma}(J^{(i)}|\phi^{(i)}, \phi, I_0)$ in \S \ref{subsec:likelihood} based on a Riemannian structure of the ODFs.

\subsection{The Shape Prior of the Atlas $f(\phi | I_0)$ and the Distribution of Random Diffeomorphisms $f(\phi^{(i)} | \phi, I_0)$ }

Adopting previous work \citep{jun_nimg_2008,Qiu:TIP10} , we discuss the construction of the shape prior (probability distribution) of the atlas, $f(\phi | I_0)$, under  the framework of large deformation diffeomorphic metric mapping (LDDMM, reviewed in Appendix A). By Property \ref{prop:linearization} in Appendix A, we can compute the prior $f(\phi |I_0)$ via $m_0$, \ie
\begin{align}
f(\phi|I_0) = f(m_0| I_0) \ ,
\end{align}
where $m_0$ is initial momentum defined in the coordinates of $I_0$ such that it uniquely determines diffeomorphic geodesic flows from $I_0$ to the estimated atlas. When $I_0$ remains fixed, the space of the initial momentum $m_0$ provides a linear representation of the nonlinear diffeomorphic shape space, $I_{atlas}$, in which linear statistical analysis can be applied. Hence, assuming $m_0$ is random, we immediately obtain a stochastic model for {\it diffeomorphic transformations} of $I_0$. More precisely, we follow the work in \citep{jun_nimg_2008,Qiu:TIP10} and make the following assumption.
\begin{assumption} \textbf{(Gaussian Assumption on $m_0$)\quad}
\label{assume:first}
$m_0$ is assumed to be a centered Gaussian random field (GRF) model where the distribution of $m_0$ is characterized by its covariance bilinear form, defined by
\begin{align*}
\Gamma_{m_0}(v, w) = E\bigl[ m_0(v) m_0(w) \bigr] \ ,
\end{align*}
where $v,w$ are vector fields in the Hilbert space of $V$ with reproducing kernel $k_V$. 
\end{assumption}
\noindent We associate $\Gamma_{m_0}$ with $k_V^{-1}$. 
The ``prior'' of $m_0$ in this case is then $\frac{1}{\mathcal{Z}}\exp{\left(-\frac{1}{2}\langle m_0, k_V m_0 \rangle_2\right)}$, where $\mathcal{Z}$ is the normalizing Gaussian constant. This leads to formally define the ``log-prior'' of $m_0$ to be
\begin{align}
\label{eqn:log-prior}
\log f(m_0 | I_0) \approx -\frac{1}{2} \langle m_0, k_V m_0 \rangle_2  \ ,
\end{align}
where we ignore the normalizing constant term $\log{\mathcal{Z}}$.

\bigskip We now consider the construction of the distribution of random diffeomorphisms, $f(\phi^{(i)} | \phi, I_0)$. Similar to the construction of the atlas shape prior, we define $f(\phi^{(i)} | \phi, I_0)$ via the corresponding initial momentum $m_0^{(i)}$ defined in the coordinates of $\phi \cdot I_0$. We also assume that $m_0^{(i)}$ is random, and therefore, we again obtain a stochastic model for {\it diffeomorphic transformations} of $I_{\atlas} \cong \phi \cdot I_0$. We make the following assumption.
\begin{assumption} \textbf{(Gaussian Assumption on $m_0^{(i)}$)\quad}
\label{assume:second}
$m_0^{(i)}$ is assumed to be a centered GRF model with its covariance as $k_V^\pi$, where $k_V^{\pi}$ is the reproducing kernel of the smooth vector field in a Hilbert space $V$.
\end{assumption}
\noindent Hence, we can define the log distribution of random diffeomorphisms as 
\begin{align}
\label{eqn:loglikelihoodrandiff}
 \log f(\phi^{(i)} | \phi, I_0) \approx -\frac{1}{2} \langle m_0^{(i)}, k_V^{\pi} m_0^{(i)} \rangle_2  \ .
\end{align} 
where as before, we ignore the normalizing constant term $\log{\mathcal{Z}}$.

\subsection{The Conditional Likelihood of the ODF Data $f_{\sigma}(J^{(i)}|\phi^{(i)}, \phi, I_0)$}
\label{subsec:likelihood}

In this section, we will derive the construction of the conditional likelihood of the ODF data $f_{\sigma}(J^{(i)}|\phi^{(i)}, \phi, I_0)$. 
From the field of {\it information geometry} \citep{Amari85}, the space of ODFs, $\p(\s)$, forms a Riemannian manifold with the {\it Fisher-Rao} metric (reviewed in Appendix B). In our study, we choose the square-root representation of the ODFs as the parameterization of the ODF Riemannian manifold, which was used recently in ODF processing and registration  \citep{Du:TMI_2011,Goh:NeuroImage2011,Cheng:MICCAI09}. The {\it square-root ODF} ($\gsODF$) is defined as
$\displaystyle
\bpsi(\s)=\sqrt{\p(\s)}$, where $\bpsi(\s)$ is assumed to be non-negative to ensure uniqueness. The space of such functions is defined as
\begin{align}
\label{eq:sphereeqn}
\bPsi=
\{\bpsi:\Do^2\rightarrow \Re^+ | \forall \s\in\Do^2, \bpsi(\s) \geq 0; \int_{\s\in\Do^2} \bpsi^2(\s) d\s=1\}.
\end{align}
We refer the interested reader to Appendix B for a more detailed description of the Riemmanian manifold $\bPsi$ lies on.
It can be shown \citep{Srivastava:CVPR07} that the Fisher-Rao metric is simply the $\bL^2$ metric, given as
\begin{align}
\label{eq:FisherRao}
\langle \bxi_j, \bxi_k \rangle_{\bpsi_i} = 
\int_{\s\in\Do^2} \bxi_j(\s) \bxi_k (\s) d\s,
\end{align}
where $\bxi_j, \bxi_k \in T_{\bpsi_i} \bPsi$ are tangent vectors at $\bpsi_i$. 
As we see from the preceding discussion, the ODF image should instead be considered as a function indexed over a unit sphere $\Do^2$ and the image volume $\bOmega\subset\Re^3$. We denote $J^{(i)}$ as $\bpsi^{(i)}(\s,x)$, $\s \in \Do^2,  x \in \bOmega$ in the remainder of the paper. Similarly, we have the atlas $I_{\atlas}=\bpsi_{\atlas}(\s,x)$, where $\bpsi_{\atlas}(\s,x)$ not only represents the mean anatomical shape characterized through the diffeomorphism but the mean ODF at each spatial location described using $\gsODF$.

Given $\phi_1^{(i)}$ and $\bpsi_{\atlas}(\s,x)$ at a specific spatial location $x$, we assume that $\bpsi^{(i)}(\s, x)$ is generated through an exponential map, \ie, 
\begin{align}
\bpsi^{(i)}(\s, x) = \exp_{\phi_1^{(i)} \cdot \bpsi_{\atlas}(\s,x)} \Big(\bxi(x) \Big),
\end{align}
where the tangent vectors $\bxi(x) \in T_{\phi_1^{(i)} \cdot \bpsi_{\atlas}(\s,x)}\bPsi$ lie in a linear space. Therefore, in order to model  conditional likelihood of the ODF $f_{\sigma}(J^{(i)}|\phi^{(i)}, \phi, I_0)$, we make the following assumption.
\begin{assumption} \textbf{(Gaussian Assumption on $\bxi$)\quad}
\label{assume:third}
$\bxi(x) \in T_{\phi_1^{(i)} \cdot \bpsi_{\atlas}(\s,x)}\bPsi$ is assumed to be a centered Gaussian Random Field on  the tangent space of $\bPsi$ at $\phi_1^{(i)} \cdot \bpsi_{\atlas}(\s,x)$. In addition, we assume that this Gaussian random field has the covariance as $\sigma^2 \Gamma_\Id$.
\end{assumption}
\noindent This assumption is based on previous works on Bayesian atlas estimation using images and shapes \citep{Qiu:TIP10, jun_nimg_2008}. The main difference here is that we assume that $\bxi(x) \in T_{\phi_1^{(i)} \cdot \bpsi_{\atlas}(\s,x)}\bPsi$ is assumed to be a centered Gaussian Random Field on  the tangent space. We choose $\Gamma_\Id$ as the identity operator to be consistent with the inner product of $\gsODF$ defined in Eq. \eqref{eq:FisherRao}.  The group action of the diffeomorphism, $\phi_1^{(i)} \cdot \bpsi_{\atlas}(\s,x)$, involves both the spatial transformation and reorientation of the ODF. Based on the derivation in our previous work \citep{Du:TMI_2011}, we define this group action as 
\begin{align}
\label{eqn:groupaction}
\phi_1^{(i)} \cdot \bpsi_{\atlas}(\s,x)  =   \sqrt{\frac{\det{\bigl(D_{(\phi_1^{(i)})^{-1}}\phi_1^{(i)} \bigr)^{-1}} }{\left\|{\bigl(D_{(\phi_1^{(i)})^{-1}}\phi_1^{(i)} \bigr)^{-1} } \s \right\|^3} } \quad
\bpsi_{\atlas} \left( \frac{(D_{(\phi_1^{(i)})^{-1}}\phi_1^{(i)} \bigr)^{-1} \s}{\|(D_{(\phi_1^{(i)})^{-1}}\phi_1^{(i)} \bigr)^{-1} \s\|}, (\phi_1^{(i)})^{-1}(x) \right) .
\end{align}
This leads to formally define the ``log-likelihood'' of $\bxi(x)$ as
\begin{align*}
-\frac{1}{2\sigma^2} \langle \bxi, \bxi \rangle_2 = -\frac{1}{2\sigma^2} \Big\| \log_{\phi_1^{(i)} \cdot \bpsi_{\atlas}(\s,x)} \bpsi^{(i)}(\s, x)  \Big\|_{\phi_1^{(i)} \cdot \bpsi_{\atlas}(\s,x)}^2 \quad .
\end{align*}
From the Gaussian assumption, we can thus write the conditional ``log-likelihood'' of $J^{(i)}$ given $I_{\atlas}$ and $\phi_1^{(i)}$ as 
\begin{align}
\label{eqn:conditionalloglikelihood}
&\log f_{\sigma}(J^{(i)}|\phi_1^{(i)}, \phi_1, I_0)  \\ \nonumber
\approx & \int_{x \in \bOmega} \Big\{ -\frac{1}{2\sigma^2} \bigg\| \log_{\phi_1^{(i)} \cdot \bpsi_{\atlas}(\s,x)} \Big(\bpsi^{(i)}(\s, x) \Big) \bigg\|_{\phi_1^{(i)} \cdot \bpsi_{\atlas}(\s,x)}^2 -\frac{\log \sigma^2}{2} \Big\} dx  \ ,
\end{align}
where as before, we ignore the normalizing Gaussian term, and $I_0$ is denoted as $\bpsi_0(\s,x)$ such that $\bpsi_{\atlas}(\s,x)=\phi_1\cdot\bpsi_0(\s,x)$.

\subsection{Expectation-Maximization Algorithm}
\label{subsec:em}

We have shown how to compute the log-likelihood shown in Eq. \eqref{eqn:loglikelihood} in \S \ref{subsec:generalframework} and \S \ref{subsec:likelihood}. In this section, we will show how we employ the Expectation-Maximization algorithm to estimate the atlas, $I_{\atlas} =\bpsi_{\atlas}(\s,x)$, for $\s \in \Do^2, x\in \bOmega$, and $\sigma^2$. From the above discussion, we first rewrite the log-likelihood function of the complete data in Eq. \eqref{eqn:loglikelihood} as 
\begin{align}
 & \log f_{\sigma}( \phi, \phi^{(1)}, \dots, \phi^{(n)}, J^{(1)},\dots J^{(n)} | I_0)  \\  \nonumber 
 \approx&   \log f_{\sigma}( m_0, m_0^{(1)}, \dots, m_0^{(n)}, \bpsi^{(1)},\dots \bpsi^{(n)} | \bpsi_0)  \\ \nonumber
\approx& -\frac{1}{2} \langle m_0, k_V m_0 \rangle_2  \\ \nonumber
&-\sum_{i=1}^n \Bigg\{ \frac{1}{2} \langle m_0^{(i)}, k_V^{\pi} m_0^{(i)} \rangle_2 +   
   \int_{x \in \bOmega} \Big\{ \frac{1}{2\sigma^2} \Big\| \log_{\phi_1^{(i)} \cdot \bpsi_{\atlas}(\s,x)} \bpsi(\s, x)  \Big\|_{\phi_1^{(i)} \cdot \bpsi_{\atlas}(\s,x)}^2 +\frac{\log \sigma^2}{2} \Big\} dx
\Bigg\} \ ,
\end{align}
where $\bpsi_{\atlas}(\s,x) = \phi_1 \cdot \bpsi_0(\s,x)$ and can be computed based on Eq. \eqref{eqn:groupaction}.

\bigskip \noindent\textbf{The E-Step.} The E-step computes the expectation of the complete data log-likelihood given the previous atlas $m_0^{\old}$ and variance ${\sigma^{2}}^{\old}$. We denote this expectation as 
$Q(m_0, \sigma^2|m_0^{\old},{\sigma^{2}}^{\old})$ given in the equation below,
\begin{align}
\label{eqn:qfun}
& Q\left(m_0, \sigma^2|m_0^{\old},{\sigma^{2}}^{\old}\right) \\ \nonumber
= &E\Bigg\{ \log f_{\sigma}( m_0, m_0^{(1)}, \dots, m_0^{(n)}, \bpsi^{(1)},\dots \bpsi^{(n)} | \bpsi_0) \Big| m_0^{\old},{\sigma^{2}}^{\old}, \bpsi^{(1)}, \cdots,  \bpsi^{(n)}, \bpsi_0 \Bigg\} \\ \nonumber
 \approx &-\frac{1}{2} \langle m_0, k_V m_0 \rangle_2 \\ \nonumber
&- \sum_{i=1}^n E\Bigg[ \frac{1}{2} \langle m_0^{(i)}, k_V^{\pi} m_0^{(i)} \rangle_2 +    \int_{x \in \bOmega} \Big\{ \frac{1}{2\sigma^2} \Big\| \log_{\phi_1^{(i)} \cdot \bpsi_{\atlas}(\s,x)} \bpsi^{(i)} (\s, x) \Big\|_{\phi_1^{(i)} \cdot \bpsi_{\atlas}(\s,x)}^2 +\frac{\log \sigma^2}{2} \Big\} dx
\Bigg] \ .
\end{align}

\bigskip\noindent\textbf{The M-Step.} The M-step generates the new atlas by maximizing the $Q$-function with respcet to $m_0$ and $\sigma^2$.   The update equation is given as  
\begin{align}
 & m_0^{\new}, {\sigma^2}^{\new} \\ \nonumber
=  &\argmax_{m_0,\sigma^2} Q\left(m_0, \sigma^2 |m_0^{\old}, {\sigma^2}^{\old} \right) \\ \nonumber
= &\argmin_{m_0,\sigma^2} \left\{\langle m_0, k_V m_0\rangle_2 
+
\sum_{i=1}^n  E\Bigg[ \int_{x \in \bOmega} \Big\{ \frac{1}{\sigma^2} \Big\| \log_{\phi_1^{(i)} \cdot \bpsi_{\atlas}(\s,x)} \bpsi^{(i)} (\s, x)  \Big\|_{\phi_1^{(i)} \cdot \bpsi_{\atlas}(\s,x)}^2 +\log \sigma^2 \Big\} dx
\Bigg] \right\}   \ ,
\end{align}
where we use the fact that the conditional expectation of $\langle m_0^{(i)}, k_V^\pi m_0^{(i)}\rangle_2 $ is constant. We solve $\sigma^2$ and $m_0$ by separating the procedure for updating $\sigma^2$ using the current value of $m_0$, and then optimizing $m_0$ using the updated value of $\sigma^2$.

Thus, we can show that it yields the following update equations (the proof is shown later in \S \ref{secsub:update}),
\begin{align}
\label{eqn:updatesigma}
&{\sigma^2}^{\new} = \frac{1}{n} \sum_{i=1}^{n}   \int_{x \in \bOmega} \Big\| \log_{\phi_1^{(i)} \cdot \bpsi_{\atlas}(\s,x)} \bpsi^{(i)}(\s, x)  \Big\|_{\phi_1^{(i)} \cdot \bpsi_{\atlas}(\s,x)}^2 dx       \ , \\
&m_0^{\new}  =  \argmin_{m_0} \left\{ \langle m_0, k_V m_0\rangle_2+ \frac{1}{\sigma^{2new}} \int_{x \in \bOmega} \alpha(x) \Big\| \log_{\overline{\bpsi}_0(\s,x)} \Big(\phi_1 \cdot \bpsi_0(\s, x) \Big) \Big\|_{\overline{\bpsi}_0(\s,x)}^2 dx   \right\}\ ,
\label{eqn:updatemomenta}
\end{align}
where $\displaystyle\alpha(x) = \sum_{i=1}^n | D\phi_1^{(i)}(x)| $ is a weighted image volume to control the contribution of the HARDI matching errors to the total cost at each voxel level. $\displaystyle|D\phi_1^{(i)}|$ is the Jacobian determinant of $\phi_1^{(i)}$. The mean ODF $\overline{\bpsi}_0(\s,x)$ is defined as the solution to the following minimization problem
 \begin{align}
 \label{eqn:I0}
 \overline{\bpsi}_0(\s, x)  = \argmin_{\bpsi\in\bPsi} \frac{1}{2}
\sum_{i=1}^n \frac{|D\phi_1^{(i)}(x)|}{\sum_{j=1}^n |D\phi_1^{(j)}(x)|}\
\Big\| \log_{\bpsi(\s,x)} \big((\phi_1^{(i)})^{-1} \cdot \bpsi^{(i)}(\s,x) \big)\Big\|_{\bpsi(\s,x)}.
 \end{align}
To compute $\overline{\bpsi}_0(\s,x)$, the weighed Karcher mean algorithm given in \citet{Goh:NeuroImage2011} is used. In addition, from \citet{Goh:NeuroImage2011}, we also know that $\overline{\bpsi}_0(\s, x) $ is the unique solution to 
 \begin{align}
\label{eqn:I01}
\frac{1}{\sum_{j=1}^n |D\phi_1^{(j)}(x)|}
\sum_{i=1}^n |D\phi_1^{(i)}(x)| \log_{\overline{\bpsi}_0(\s, x)}
\left((\phi_1^{(i)})^{-1} \cdot \bpsi^{(i)}(\s,x)\right)=\mathbf{0}.
 \end{align}
 The variational problem listed in Eq. \eqref{eqn:updatemomenta} is referred as ``modified LDDMM-ODF mapping", where the weight $\alpha$ is introduced. We now present the steps involved in each iteration in Algorithm \ref{alg:em}. 
\begin{algorithm}
\caption{\label{alg:em}\bf (The EM Algorithm for the HARDI Atlas Generation)}
We initialize $m_0=0$. Thus, the hyperatlas $\bpsi_0$ is considered as the initial atlas. 
\begin{enumerate}
\item Apply the LDDMM-ODF mapping algorithm \citep{Du:TMI_2011} to register the current atlas to each individual HARDI dataset, which yields $m_0^{(i)}$ and $\phi_t^{(i)}$.
\item Compute $\overline{\bpsi}_0 $ according to Eq. \eqref{eqn:I0} using the weighted Karcher mean algorithm given in \citet{Goh:NeuroImage2011}.
\item Update $\sigma^2$ according to Eq. \eqref{eqn:updatesigma}.
\item Estimate $\bpsi_{\atlas} = \phi_1 \cdot \bpsi_0$, where $\phi_t$ is found by applying 
the modified LDDMM-ODF mapping algorithm as given in Eq. \eqref{eqn:updatemomenta}.
\end{enumerate}
The above computation is repeated until the atlas converges.
\end{algorithm}

\subsection{Derivation of update equations of  $\sigma^2$ and $m_0$ in EM}
\label{secsub:update}

We now derive Eqs. \eqref{eqn:updatesigma} and \eqref{eqn:updatemomenta} from ${\it {Q}}$-function in Eq. \eqref{eqn:qfun} 
for updating values of $\sigma^2$ and $m_0$. It is straightforward to obtain $\sigma^2$ by taking the derivative of $Q\left(m_0, \sigma^2 | m_0^{\old}, {\sigma^2}^{\old}\right)$ with respect to $\sigma^2$ and setting it to zero. 

For updating $m_0$, let $y=\left(\phi_1^{(i)}\right)^{-1}(x)$. By the change of variables strategy, we have 
\begin{align}
& \int_{x \in \bOmega} \Big\| \log_{\phi_1^{(i)} \cdot \bpsi_{\atlas}(\s,x)} \Big(\bpsi^{(i)} (\s, x) \Big) \Big\|_{\phi_1^{(i)} \cdot \bpsi_{\atlas}(\s,x)}^2  dx  \\ \nonumber
 = & \int_{y \in \bOmega} \Big\| \log_{ \bpsi_{\atlas}(\s,y)} \Big((\phi_1^{(i)})^{-1} \cdot \bpsi^{(i)} (\s, y) \Big) \Big\|_{\bpsi_{\atlas}(\s,y)}^2  |D\phi_1^{(i)}(y)|dy \ .
\end{align}
Therefore, we can then rewrite
{\allowdisplaybreaks
\begin{align*}
&\sum_{i=1}^n E\Bigg[ \int_{x \in \bOmega} \Big\{ \frac{1}{2\sigma^2} \Big\| \log_{\phi_1^{(i)} \cdot \bpsi_{\atlas}(\s,x)} \bpsi^{(i)} (\s, x) \Big\|_{\phi_1^{(i)} \cdot \bpsi_{\atlas}(\s,x)}^2 \Big\} dx\Bigg]\\
=&\sum_{i=1}^n E\Bigg[ \int_{y \in \bOmega} 
\frac{1}{2\sigma^2}
\Big\| \log_{ \bpsi_{\atlas}(\s,y)} \Big((\phi_1^{(i)})^{-1} \cdot \bpsi^{(i)} (\s, y) \Big) \Big\|_{\bpsi_{\atlas}(\s,y)}^2  |D\phi_1^{(i)}(y)|dy\Bigg]\\
=&\int_{y \in \bOmega} 
\frac{1}{2\sigma^2}
\sum_{i=1}^n E\Bigg[  \Big\| \log_{ \bpsi_{\atlas}(\s,y)} \Big((\phi_1^{(i)})^{-1}\cdot \bpsi^{(i)} (\s, y) \Big) \Big\|_{\bpsi_{\atlas}(\s,y)}^2  |D\phi_1^{(i)}(y)| \Bigg] dy\\
\overset{\text{(a)}}\approx&\int_{y \in \bOmega} 
\frac{1}{2\sigma^2}
 \sum_{i=1}^n E\Bigg[  \Big\| \log_{ \overline{\bpsi}_0(\s,y)} \Big((\phi_1^{(i)})^{-1} \cdot \bpsi^{(i)} (\s, y) \Big) - \log_{ \overline{\bpsi}_0(\s,y)} \Big(  \bpsi_{\atlas}(\s,y)\Big)
\Big\|_{\overline{\bpsi}_0(\s,y)}^2  |D\phi_1^{(i)}(y)| \Bigg]   dy\\
=&\int_{y \in \bOmega}  
\frac{1}{2\sigma^2}
\sum_{i=1}^n E\Bigg[  \Big\{ \Big\| \log_{ \overline{\bpsi}_0(\s,y)} \Big((\phi_1^{(i)})^{-1} \cdot \bpsi^{(i)} (\s, y) \Big) \Big\|_{\overline{\bpsi}_0(\s,y)}^2 + \Big\|\log_{ \overline{\bpsi}_0(\s,y)} \Big(  \bpsi_{\atlas}(\s,y)\Big)
\Big\|_{\overline{\bpsi}_0(\s,y)}^2  \\ \nonumber 
& -2 \Big\langle \log_{ \overline{\bpsi}_0(\s,y)} \Big((\phi_1^{(i)})^{-1} \cdot \bpsi^{(i)} (\s, y) \Big), \log_{ \overline{\bpsi}_0(\s,y)} \Big(  \bpsi_{\atlas}(\s,y)\Big) \Big\rangle_{\overline{\bpsi}_0(\s,y)} \Big\}  |D\phi_1^{(i)}(y)| \Bigg] dy
\end{align*}}
where (a) is the first order approximation of $\Big\| \log_{ \bpsi_{\atlas}(\s,y)} \Big((\phi_1^{(i)})^{-1}\cdot \bpsi^{(i)} (\s, y) \Big) \Big\|_{\bpsi_{\atlas}(\s,y)}^2$.

As the direct consequence of the Karcher mean definition of $\overline{\bpsi}_0(\s,y)$ in Eq. \eqref{eqn:I0}, and more precisely Eq. \eqref{eqn:I01}, $\sum_{i=1}^n |D\phi_1^{(i)}(x)| \log_{\overline{\bpsi}_0(\s, x)}
\left((\phi_1^{(i)})^{-1} \cdot \bpsi^{(i)}(\s,x)\right)=\mathbf{0}$, the above cross item is equal to zero. Therefore, we get 
\begin{align*}
\int_{y \in \bOmega}
\frac{1}{2\sigma^2}
\sum_{i=1}^n& 
E\Bigg[  \Big\{ \Big\| \log_{ \overline{\bpsi}_0(\s,y)} \Big((\phi_1^{(i)})^{-1} \cdot \bpsi^{(i)} (\s, y) \Big) \Big\|_{\overline{\bpsi}_0(\s,y)}^2 \\ &+ \Big\|\log_{ \overline{\bpsi}_0(\s,y)} \Big(  \bpsi_{\atlas}(\s,y)\Big)
\Big\|_{\overline{\bpsi}_0(\s,y)}^2 \Big\} |D\phi_1^{(i)}(y)| \Bigg] dy.
\end{align*}
Since the first item in the above equation is independent of $m_0$, we have
\begin{align*}
m_0^{\new} & =  \argmin_{m_0}  \langle m_0, k_V m_0\rangle_2 +\frac{1}{\sigma^{2new}} \int_{y \in \bOmega} \alpha(y) \Big\| \log_{\overline{\bpsi}_0(\s,y)} \Big(\phi_1 \cdot \bpsi_0(\s, y) \Big) \Big\|_{\overline{\bpsi}_0(\s,y)}^2 dy   \ ,
\end{align*}
where $\alpha(y) = \sum_{i=1}^n |D\phi_1^{(i)}(y)| $. By changing $y$ by $x$, we obtain Eq. \eqref{eqn:updatesigma}.


\section{Results}
\label{sec:experiments}

In this section, we demonstrate the performance of the probabilistic HARDI atlas generation algorithm proposed on real human data. In \S \ref{sec:hardiwholeatlas}, we show the HARDI atlas based on $94$ healthy adults. \S \ref{sec:hardiatlascon} empirically examines the convergence of the HARDI atlas estimation procedure and studies the effects of the choice of the hyperatlas, which is used as the initial atlas in Algorithm \ref{alg:em}, on the final estimated atlas. \S \ref{sec:atlasaging} shows the estimated atlases across different age groups. Finally, \S \ref{subsec:comparison} compares our proposed algorithm to an existing algoritim in \citet{Bloy2011}.

\bigskip \noindent\textbf{Subjects and Image Acquisition:}
$94$ participants were recruited through advertisements posted at the National University of Singapore (NUS). $38$ males and $56$ females ranged from $22$ to $71$ years old (mean $\pm$ standard deviation (SD): $42.5 \pm  13.9$ years) participated in the study. A health screening questionnaire along with informed consent approved by the NUS Institutional Review Board was acquired from each participant. Any participant with a history of psychological, neurological disorder or surgical implantation was excluded from the study. A Mini Mental Status Examination (MMSE) was administered to each participant to rule out possible cognitive impairments. All participants had the MMSE score greater than $26$. 

Every participant underwent magnetic resonance imaging scans that were performed on a $3$T Siemens Magnetom Trio Tim scanner using a $32$-channel head coil at Clinical Imaging Research Center at the NUS. The image protocols were: (i) isotropic high angular resolution diffusion imaging (single-shot echo-planar sequence; $48$ slices of $3$mm thickness; with no inter-slice gaps; matrix: $96 \times 96$; field of view: $ 256 \times 256$mm; repetition time: $6800$ ms; echo time: $85$ ms; flip angle: $90^\circ$; $91$  diffusion weighted images (DWIs) with $b=1150$ s/mm$^2$, $11$ baseline images without diffusion weighting); (ii) isotropic T$2$-weighted imaging protocol (spin echo sequence; $48$ slices with $3$ mm slice thickness; no inter-slice gaps; matrix: $96 \times 96$; field of view: $256 \times 256$ mm; repetition time: $2600$ ms; echo time: $99$ ms; flip angle: $150^\circ$).

\bigskip \noindent\textbf{HARDI Preprocessing:}
DWIs of each subject were first corrected for motion and eddy current distortions using affine transformation to the image without diffusion weighting. Within-subject, we followed the procedure detailed in \citet{Huang:MRI} to correct geometric distortion of the DWIs due to b$0$-susceptibility differences over the brain. Briefly reviewing, the T$2$-weighted image was considered as the anatomical reference. The deformation that carried the baseline image without diffusion weighting to the T$2$-weighted image characterized the geometric distortion of the DWI. For this, intra-subject registration was first performed using FLIRT \citep{jenkinson:FLIRT} to remove linear transformation (rotation and translation) between the diffusion weighted images and T$2$-weighted image. Then, LDDMM \citep{Du:SD} sought the optimal nonlinear transformation that deformed the baseline image without the diffusion weighting to the T$2$-weighted image. This diffeomorphic transformation was then applied to every diffusion weighted image in order to correct the nonlinear geometric distortion.  
Existing literature \citep{Tao:Miccai06,Dhollander:CDMRI2010} have proposed different ways of reorienting the diffusion gradients. In this paper, the diffusion gradients are reoriented using the method proposed in \citet{Dhollander:CDMRI2010}. Briefly speaking, if $\phi$ is the diffeomorphism, then the local affine transformation $A_{x}$ at spatial coordinates $x$ is defined as the Jacobian matrix of $\phi$ evaluated at $x$. If $\bg_i$ is the $i^{th}$ diffusion gradient, then the reoriented diffusion gradient after the affine transformation $A_{x}$ is simply $\frac{A_{x}^{-T}\bg_i}{\|A_{x}^{-T}\bg_i\|}$. Finally, we estimated the ODFs using the approach considering the solid angle constraint based on DWI images proposed in \citet{Aganj:MRM10}.

\subsection{HARDI Atlas Generation}
\label{sec:hardiwholeatlas}

To initialize the HARDI atlas generation process,  we chose the HARDI dataset of one participant (male, $43$ years old) as hyperatlas and assumed $m_0=0$ such that the hyperatlas was used as the initial atlas. We then followed Algorithm \ref{alg:em} and ten iterations were repeated. Notice that $k_V$ associated with the covariance of $m_0$ and $k_V^\pi$ associated with the covariance of $m_0^{(i)}$ were assumed to be known and predetermined. Since we were dealing with vector fields in $\Re^3$, the kernel of $V$ is a matrix kernel operator in order to get a proper definition. Making an abuse of notation, we defined $k_V$ and $k_V^\pi$ respectively as $k_V \Id_{3 \times 3}$ and $k_V^\pi \Id_{3 \times 3}$,  where $\Id_{3 \times 3}$ is a $3 \times 3$ identity matrix and $k_V$ and $k_V^\pi$ are scalars.  In particular, we assumed that $k_V$ and $k_V^\pi$ are Gaussian with kernel sizes of $\sigma_V$ and $\sigma_{V^\pi}$.
Since $\sigma_V$ determines the smoothness level of the mapping from the hyperatlas to the blur  $\overline{\bpsi}_0(\s, x)$ whereas $\sigma_{V^\pi}$ determines that from the sharp atlas to individual HARDI datasets, $\sigma_V$ should be greater than $\sigma_{V^\pi}$. We experimentally determined $\sigma_{V^\pi}=5$ and $\sigma_V=8$.

\begin{figure}[htb]
\centering
\includegraphics[width=0.95\linewidth]{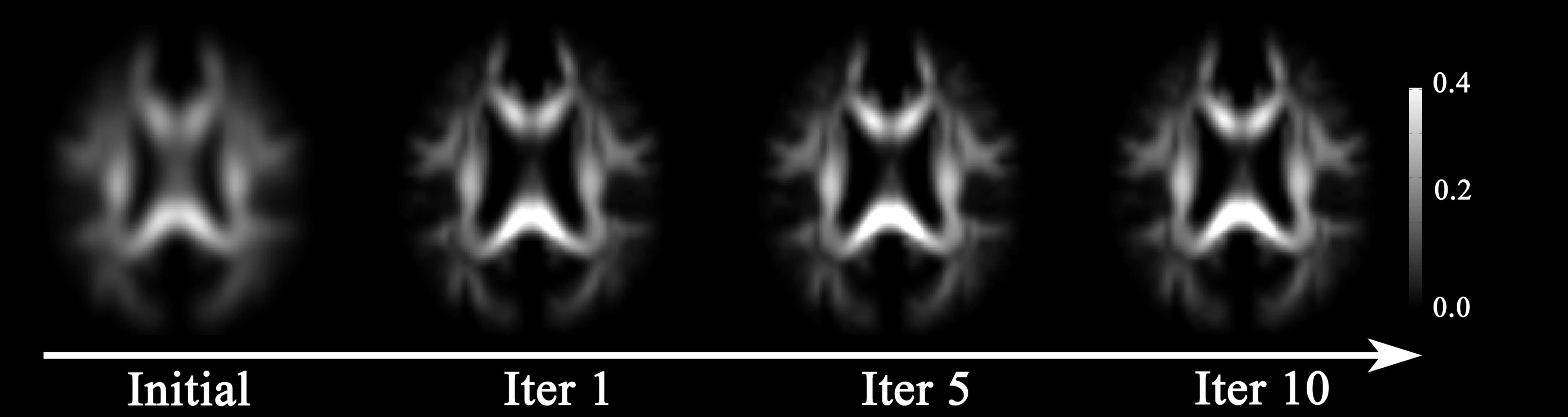}
\caption{The evolution of $\overline{\bpsi}_0(\s, x)$ over the optimization of the atlas estimation. Panels from left to right show $\overline{\bpsi}_0(\s, x)$ before the optimization, at the first, fifth, and tenth iterations, respectively. The intensity indicates the $\gsODF$ metric of each voxel with respect to the spherical ODF. The larger the value, the more anisotropic the ODF is.}
\label{fig:averagepsi}
\end{figure}

\begin{figure}[htb]
\centering
\includegraphics[width=0.95\linewidth]{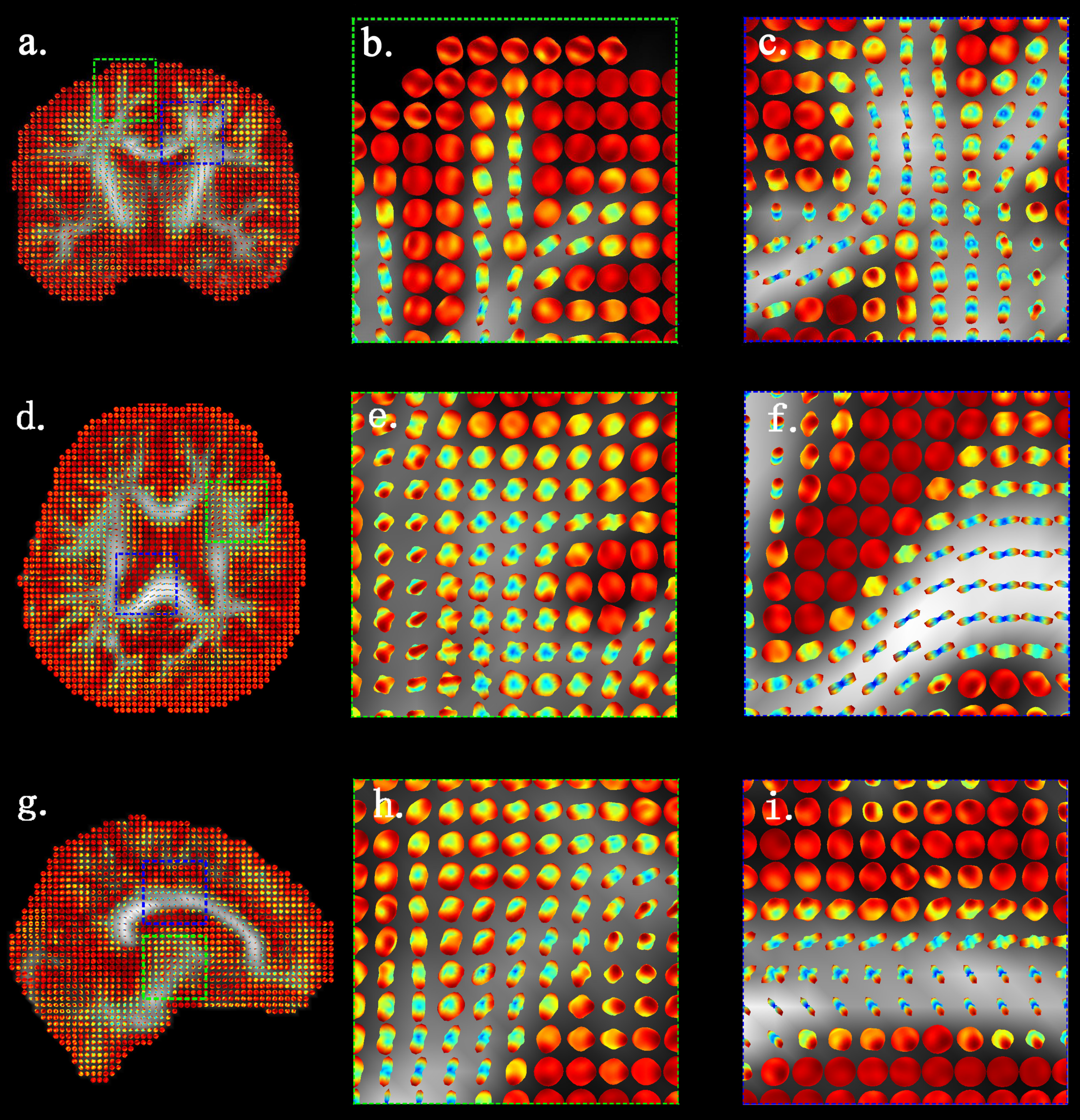}
\caption{Illustration of the branching and crossing bundles in the estimated atlases over the entire population group. Panels (a,d,g) show the ODF field in the coronal, axial, and sagittal views. In each row, the second and third panels show two zoom-in regions for branching and crossing bundles corresponding to the anatomy on the first panel.}
\label{Fig:AtlasBranchingCrossing}
\end{figure}

Figure \ref{fig:averagepsi} shows the evolution of $\overline{\bpsi}_0(\s, x)$ over the iterations of the EM algorithm. As seen in Figure \ref{fig:averagepsi}, the white matter anatomy of $\overline{\bpsi}_0(\s, x)$ was blur at the initial estimate and became sharper as more iterations were run. The computational time for each LDDMM-ODF mapping was about $30$ minutes. Figure \ref{fig:estimatedatlas} illustrates the atlas estimated from the $94$ adults' HARDI datasets after ten iterations. Panels (a-c) shows the coronal view of the atlas, while panels (d-f) and (g-i) respectively illustrate the axial and sagittal views of the atlas. Figure \ref{Fig:AtlasBranchingCrossing} shows the branching and crossing bundles in the estimated atlases over the entire population group, suggesting that the atlas preserves the anatomical details of the white matter. 

\subsection{Convergence and Effects of Hyperatlas Choice of the HARDI Atlas Estimation}
\label{sec:hardiatlascon}

\begin{figure}[!htb]
\centering
\includegraphics[width=0.5\linewidth]{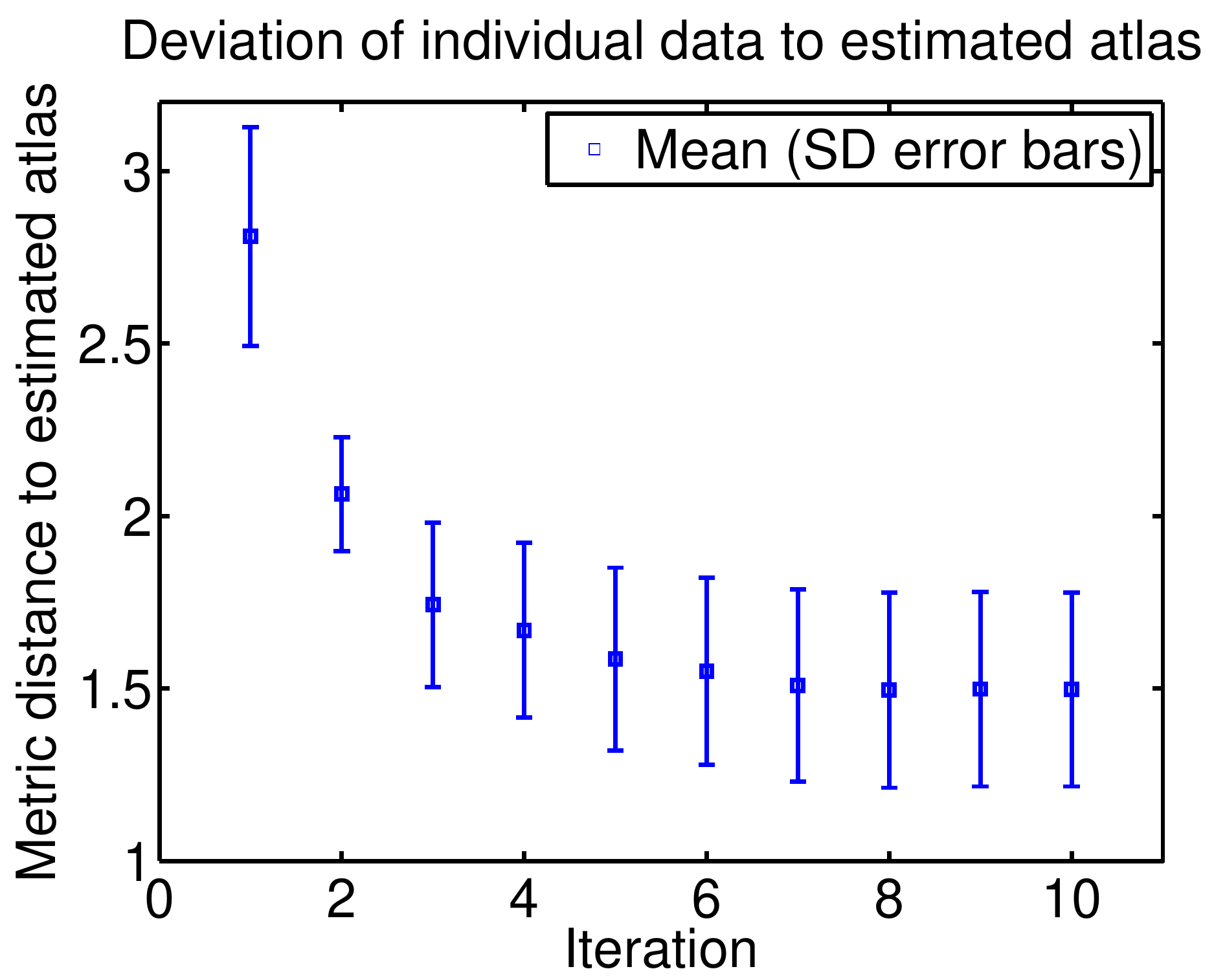}
\caption{The evolution of the average diffeomorphic metric between individual subjects and the estimated atlas, with the standard deviation shown by the error bars.}
\label{fig:optdiffeometric}
\end{figure}

In this section, we empirically demonstrate the convergence of the average diffeomorphic metric of individual subjects when referenced to the estimated atlas. This is measured using the square root of the inner product of the initial momentum. Figure \ref{fig:optdiffeometric} shows the evolution of the average diffeomorphic metric of individual subjects referenced to the estimated atlas as well as the standard deviation across the subjects. From Figure \ref{fig:optdiffeometric}, we see that the average diffeomorphic metric changed less than $5\%$ after two iterations. 


Next, we study the effects of the hyperatlas choice on the estimated atlas. In the Bayesian modeling for the HARDI atlas generation presented here, the hyperatlas $\bpsi_0$ is assumed to be known and fixed. In addition, the hyperatlas is used as the initialization for the atlas in the EM algorithm. Therefore, the anatomy of the estimated atlas can be dependent on the choice of the hyperatlas. In this section, we demonstrate the influence due to the hyperatlas.

\begin{figure}[!htb]
\centering
\includegraphics[width=0.95\linewidth]{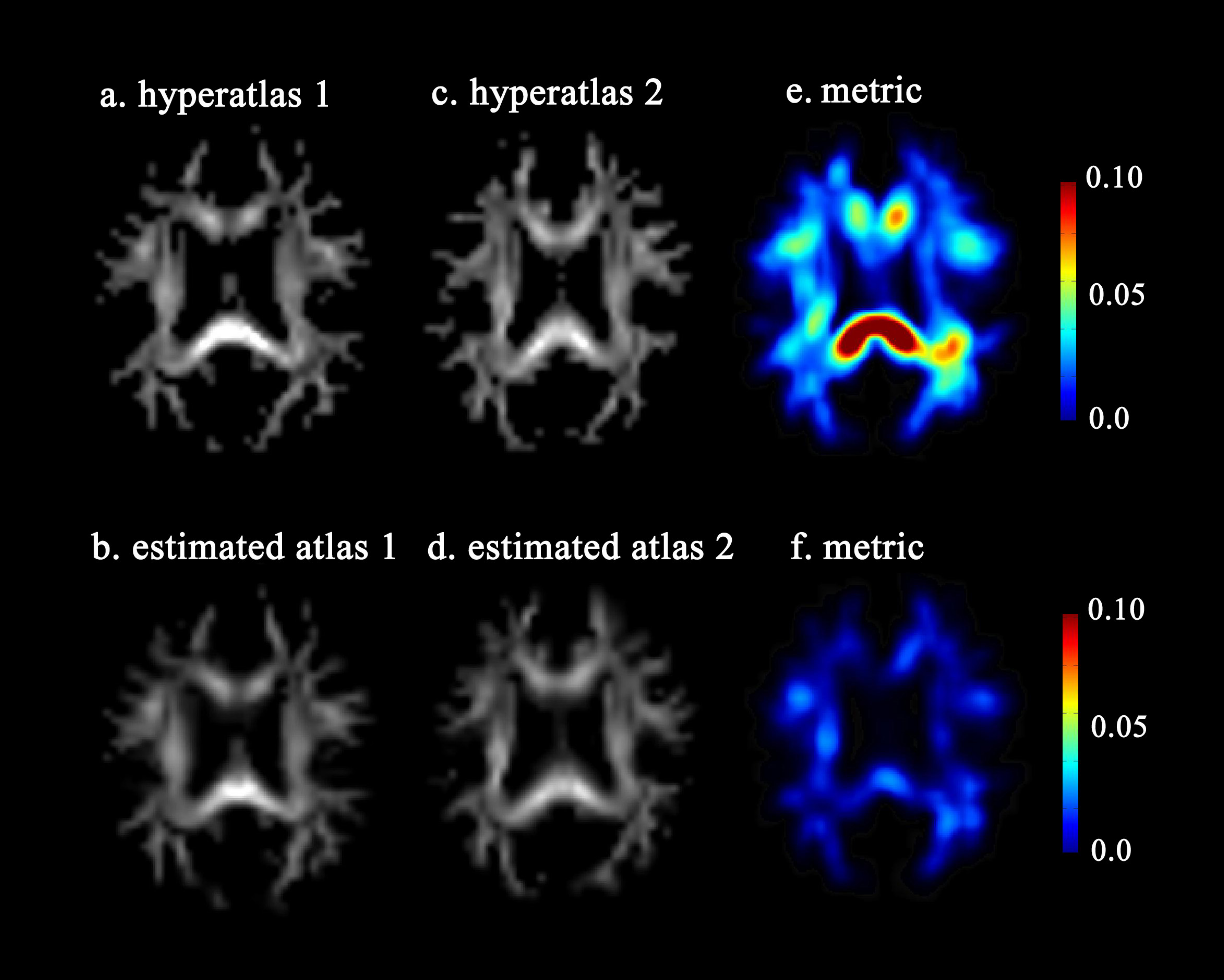}
\caption{Influences of the hyperatlas on the estimated atlas.  
Two HARDI datasets (panels (a, c)) were respectively used as the hyperatlas in the Bayesian atlas estimation, which generated the atlases 
shown in panels (b, d). Panel (e) shows the $\gsODF$ metric square between the two hyperatlases on (a, c), while panel (f) shows that between the atlases on (b, d).}
\label{fig:hyperatlas}
\end{figure}

\begin{figure}[!htb]
\centering
\includegraphics[width=0.5\linewidth]{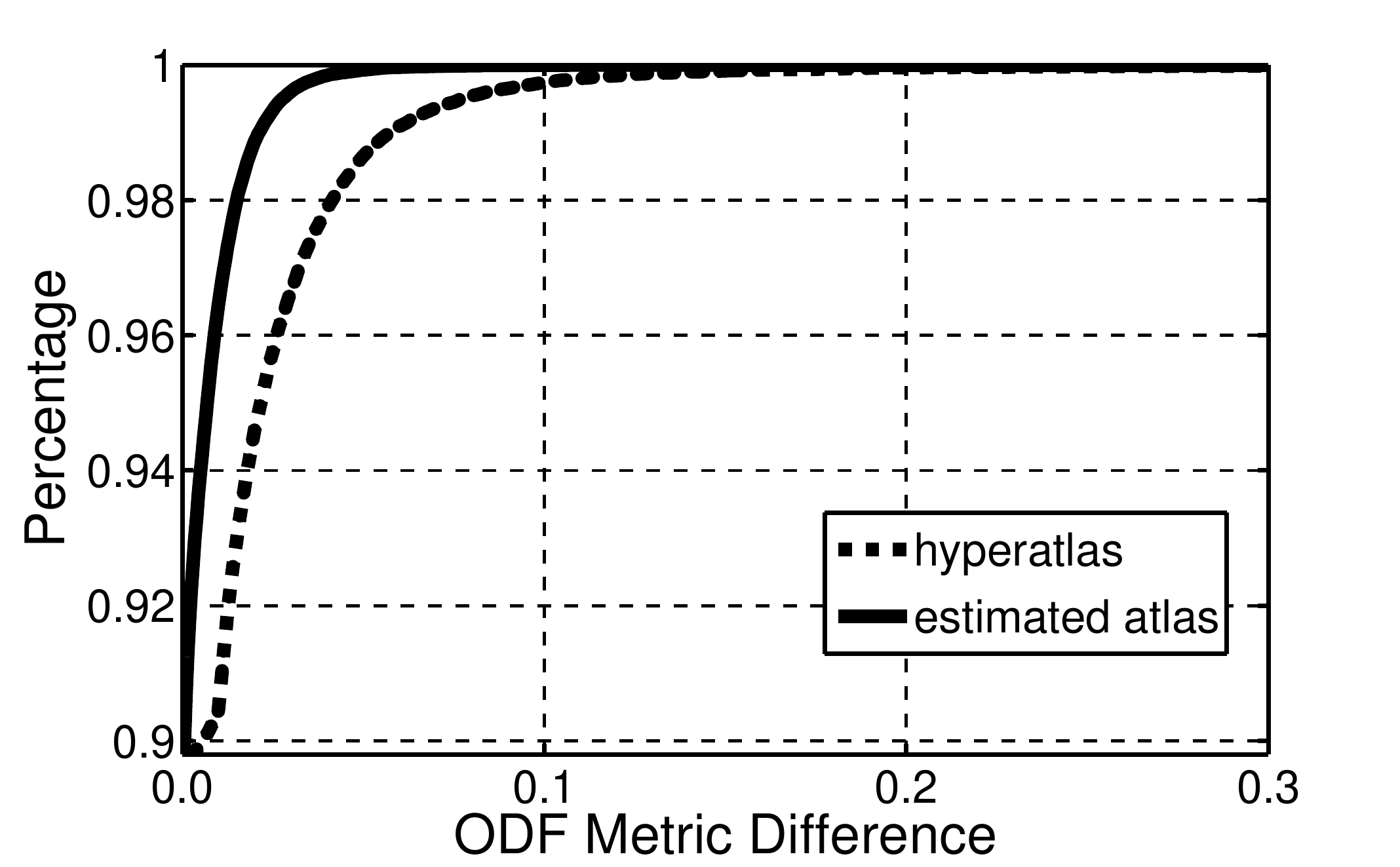}
\caption{The cumulative distributions of the $\gsODF$ metric square between the two hyperatlases (Figure \ref{fig:hyperatlas}  (a, c)) and between the two estimated atlases (Figure \ref{fig:hyperatlas}  (b, d)) are respectively shown in the dashed and solid lines.}
\label{fig:hyperatlas_CC}
\end{figure}

We repeated the atlas estimation procedure when two different HARDI datasets, shown in Figure \ref{fig:hyperatlas} (a, c), are respectively used as the hyperatlas. In this experiment, instead of using the entire dataset of $94$ adults, only ten HARDI datasets were chosen from our sample pool as the observables, $\bpsi^{(i)}$, $i=1,2, \dots, 10$. Figure \ref{fig:hyperatlas} (b, d) show the estimated HARDI atlases obtained from the hyperatlases shown in Figure \ref{fig:hyperatlas} (a, c), respectively. 
As seen in Figure \ref{fig:hyperatlas} (e), differences between the two hyperatlases are large in terms of the $\gsODF$ metric square even in major white matter bundles (e.g., corpus callosum, external capsule). Nevertheless, Figure \ref{fig:hyperatlas} (f), which shows the $\gsODF$ metric square between the estimated two atlases, illustrates that they are similar. A two-sample Kolmogorov-Smirnov test revealed that the cumulative distribution of the $\gsODF$ metric square as shown in Figure \ref{fig:hyperatlas_CC} between the two estimated atlases (Figure \ref{fig:hyperatlas}  (b, d)) is significantly greater than that between the two hyperatlases (Figure \ref{fig:hyperatlas}  (a, c)) ($p<0.001$), which indicates that more voxels with small $\gsODF$ between the two estimated atlases when compared to those between the two hyperatlases. This result suggests that the choice of the hyperatlas has minimal effects on the resulting estimated atlas.

\clearpage
\subsection{Aging HARDI atlases}
\label{sec:atlasaging}

In this section, we performed our HARDI atlas generation process on two different age groups, young and old adults, and demonstrated that the estimated atlas of each specific age group exhibits characteristics of the group that are in line with what is reported in current literature.

We selected a subset of the dataset and divided them into two groups. In the young adults group, there were $21$ subjects ($8$ males and $13$ females) ranging from $22$ to $39$ years old (mean $\pm$ standard deviation (SD): $27.6 \pm 4.28$ years);  In the old adults group, there were also $21$ subjects ($9$ males and $12$ females) ranging from $55$ to $71$ years old (mean $\pm$ standard deviation (SD): $61.90 \pm 3.81$ years). Next, we choose one subject (male, $24$ years old) as the hyperatlas for the young adults group and another subject (male, $71$ years old) for the old adults group, and performed the proposed atlas generation algorithm shown in Algorithm \ref{alg:em} for each of the two groups.

\begin{figure}[htb!]
\centering
\includegraphics[width=0.7\linewidth]{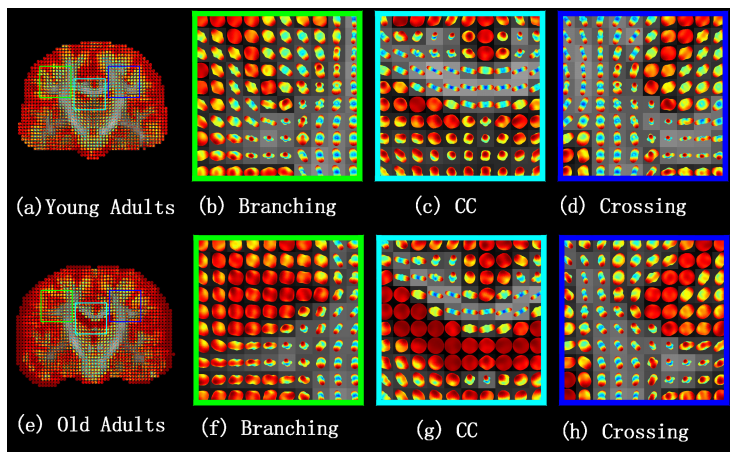}
\caption{Comparison of HARDI atlases respectively generated from young and old adults. In each row, the last three columns show three zoom-in regions for branching and crossing bundles corresponding to the anatomy given on the first panel.}
\label{Fig:AA}
\end{figure}

In Figure \ref{Fig:AA}, three regions of interest are selected for comparison between the atlases for young and old adults groups. For the regions of the corpus callosum and ventricles in panels (c, g), the most obvious aging effect observed is the bending of the corpus callosum due to the enlargement of ventricles, together with the thinning of the corpus callosum, which is consistent with previous findings in \citep{Hinkle:preprint2012,Fletcher:MFCA2011,Sullivan:CC2002}. For the region of the branching fibers, panels (b, f)  show that there are more branches in the atlas of young adults group than those in the one of old adults group. The similar effect is also observed in the region of the crossing fibers in panels (d, h).
A detailed comparison of the ODF shape explains that the anisotropy for the ODFs declines with advancing age due to the fact that axons' distribution becomes more uniform as age increases. This ODF shape differences could be due to the breakdown of the myelin sheath with aging and increases in extracellular fluid and transverse diffusivity as suggested in \citet{Michael_DTI_Aging_Review}.

\subsection{Comparison with existing method}
\label{subsec:comparison}

In this section, we compared our proposed method with the one proposed in \citet{Bloy2011}. In the rest of this section, we referred the atlas generated from our proposed method as \emph{Bayesian} atlas, and the one from \citet{Bloy2011} as \emph{averaged} atlas. While the code used in \citet{Bloy2011} is not publicly available, we manage to adapt it into the same LDDMM framework as our proposed method. To implement the ODF-based registration algorithm in \citet{Bloy2011}, we minimized the mean square error (MSE) of the spherical harmonic coefficients (SHC) of ODFs between the warped atlas and subjects, and then applied the finite strain scheme, which only keeps the rotation part of the local Jacobian field, to reorientate the ODFs. To generate an average atlas for the dataset, we first selected the same subject as the hyperatlas, and warped each subject into the hyperatlas space using by the registration method we describe above. Finally, we generated the average atlas by averaging the SHC across all the warped subjects. For a fair comparison, we kept all other conditions the same for the generation of both \emph{Bayesian} and \emph{averaged} atlases, and conducted the experiments by selecting the same hyperatlas for the entire dataset. 

\begin{figure}[htb]
\centering
\includegraphics[width=0.7\linewidth]{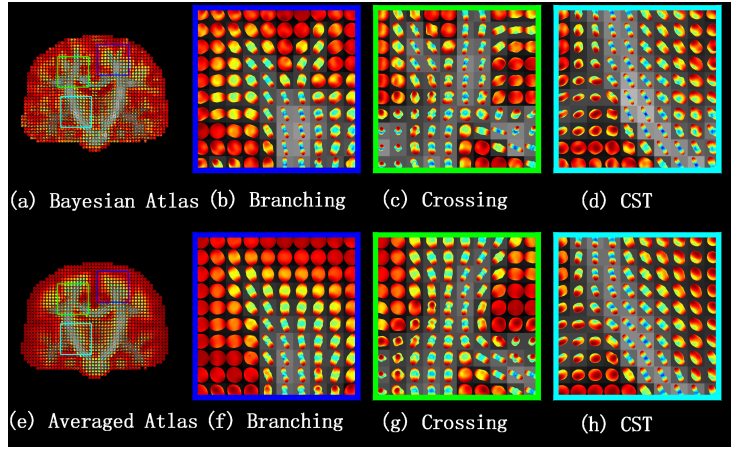}
\caption{Comparison between Bayesian and averaged atlases.  In each row, the last three columns show three zoom-in regions for branching and crossing bundles corresponding to the anatomy given on the first panel.}
\label{Fig:CBAA}
\end{figure}

As shown in Figure \ref{Fig:CBAA}, the ODFs in the \emph{Bayesian} atlas is generally much sharper than those in the averaged atlas. Moreover, as demonstrated in panels (b, f), some small branches can only be revealed in the \emph{Bayesian} atlas, while they cannot be found in the \emph{averaged} atlas due to the averaging process. Furthermore, in the region of crossing fibers  shown in panels (c, g), the \emph{Bayesian} atlas preserved more details than the \emph{averaged} atlas. However, there was not much difference in the main fiber tract as illustrated in panels (d, h).

\section{Conclusion}

In this paper, we present a Bayesian model to estimate the white matter atlas from observed HARDI datasets under the LDDMM framework. To the best of our knowledge, this is the first probabilistic approach for the HARDI atlas generation. In this work, we construct the ODF likelihood function based on its Riemannian structure. In particular, we employ the square root parameterization of the ODF Riemannian manifold such that the logarithmic and exponential maps are in closed forms. This facilitates the construction of the ODF likelihood through the tangent vector of the ODF, \ie logarithmic map, lying in a linear space where linear statistical models can be applied. We further derive the EM algorithm for solving this atlas generation problem. We empirically demonstrate the convergence of this algorithm in terms of both diffeomorphic metric and the ODF metric and show that the estimated atlas has little influence from the hyperatlas. The comparison with the existing algorighm in \citet{Bloy2011} showed that our algorithm preserves sharpness of cross and branch filbers. Hence, 
this atlas generated using our approach will be valuable for population-based studies based on HARDI.

\section*{Acknowledgments}
\addcontentsline{toc}{section}{Acknowledgments}
The work reported here was supported by grants A*STAR SICS-09/1/1/001,  a centre grant from the National Medical Research Council (NMRC/CG/NUHS/2010), the Young Investigator Award at National University of Singapore (NUSYIA FY10 P07), and National University of Singapore MOE AcRF Tier 1.


\section*{Appendix A: The Shape Prior of the Atlas $f(\phi | I_0)$ and the Distribution of Random Diffeomorphisms $f(\phi^{(i)} | \phi, I_0)$ }
\label{app:prior}
\renewcommand{\theequation}{\thesection-\arabic{equation}}  

Adopting previous work \citep{jun_nimg_2008,Qiu:TIP10},  we will first derive the construction of the shape prior (probability distribution) of the atlas, $f(\phi | I_0)$, and the distribution of random diffeomorphisms $f(\phi^{(i)} | \phi, I_0)$, by first reviewing the framework of large deformation diffeomorphic metric mapping (LDDMM). We will then show how one would define the shape prior via the initial momentum $m_0$ in LDDMM. Finally, using the same construction as in the case of the shape prior, we illustrate the construction of the distribution of the random diffeomorphisms  $f(\phi^{(i)} | \phi, I_0)$.

In LDDMM, we assume that the atlas $I_{\atlas}$ is constructed as an orbit of $I_0$ under the group of diffeomorphic transformations $\G$, \ie ${\mathcal{I}}_{\atlas} = \G \cdot I_0$. The diffeomorphic transformations are introduced as transformations of the coordinates on the background space $\Omega\subset\Real^3$, \ie $\G: \Omega \rightarrow \Omega$. One approach, proposed by \citet{grenander-miller-1998} and adopted in this paper, is to construct diffeomorphisms $\phi_t \in \G$ as a geodesic flow generated via ordinary differential equations (ODEs), where $\phi_t, t \in [0,1]$ obeys the following equation,
\begin{align}
\dot{\phi_t} = v_t(\phi_t), \quad  \phi_0 = \Id , \quad t \in [0,1],
\label{eqn:forward-flow-equation}
\end{align}
where $\Id$ denotes the identity map and $v_t$ are the associated velocity vector fields. The vector fields $v_t$ are constrained to be sufficiently smooth, so that Eq. \eqref{eqn:forward-flow-equation} is integrable and generates diffeomorphic transformations over finite time. The smoothness is ensured by forcing $v_t$ to lie in a smooth Hilbert space ($V$, $\| \cdot \|_V$) with $s$-derivatives having finite integral square and zero boundary \citep{DuGrMi1998,tro98}. In our case, we model $V$ as a reproducing kernel Hilbert space with a linear operator $L$ associated with the norm square $\| u \|_V^2 = \langle Lu, u\rangle_2$, where $\langle\cdot, \cdot\rangle_2$ denotes the $\bL^2$ inner product. The group of diffeomorphisms $ {\mathcal G}(V)$ are the solutions of Eq. \eqref{eqn:forward-flow-equation} with the vector fields satisfying $\int_0^1 \| v_t \|_V dt <\infty$.  Thus, this geodesic $\phi_t, t\in [0,1]$ which lies in the manifold of diffeomorphisms generates $I_{\atlas}$ from $I_0$, is defined as
\begin{align*}
\phi_0= \Id,\quad \phi_1 \cdot I_{0}= I_{\atlas}.
\end{align*}
The length of this geodesic is then defined as the Riemannian length of $\phi_t$, computed as the integral of the norm of the vector field $\| v_t \|_V$ associated with $\phi_t$. Alternatively, by using the duality isometry in Hilbert spaces, one can show that this geodesic length can be equivalently expressed in terms of the momentum $m_t$. $m_t$ is defined as a linear transformation of $v_t$ through kernel $k_V=L^{-1}$ associated with the reproducing kernel Hilbert space $V$. More precisely, $k_V$ maps $v_t$ to $m_t$, \ie $k_V: v_t \rightarrow m_t=k_V^{-1}v_t$. Therefore, for any $u\in V$, $\langle m_t, u\rangle_2 = \langle k_V^{-1}v_t,u\rangle_2$, where $\langle\cdot, \cdot\rangle_2$ denote the $\bL^2$ inner product. One can prove that $m_t$ satisfies the following property at all times \citep{miller-trouve-younes-2003-geodesic-shooting}.
\\
\noindent\textbf{Conservation Law of Momentum.} {\it For all $u \in V$,
\begin{align}
\label{eq:conmom}
\langle m_t, u\rangle _2 = \langle m_0, (D\phi_t)^{-1} u(\phi_t)\rangle _2.
\end{align}}
Eq. \eqref{eq:conmom} uniquely specifies $m_t$ as a linear form on $V$, given the initial momentum $m_0$ and the evolving diffeomorphism $\phi_t$. We see that by making a change of variables and obtain the following expression relating $m_t$ to the initial momentum $m_0$ and the geodesic $\phi_t$ connecting $I_0$ and $I_{\atlas}$,
\begin{equation}
m_t= |D \phi_t^{-1} | (D\phi_t^{-1} )^\top m_0 \circ \phi_t^{-1}.
\label{conservation-law-equation}
\end{equation}
As a direct consequence of this property, given the initial momentum $m_0$, one can generate a unique time-dependent diffeomorphic transformation. As a result of the preceding discussion, the following property holds true.
\begin{property}
\label{prop:linearization}
When $I_0$ remains fixed, the space of the initial momentum $m_0$ provides a linear representation of the nonlinear diffeomorphic shape space, $I_{\atlas}$, in which linear statistical analysis can be applied.
\end{property}

\section*{Appendix B: Riemannian Manifold of Square-Root ODF}
\label{app:riemanODF}


The ODF is a PDF defined on a unit sphere $\Do^2$ and its space is defined as
\begin{align*}
\P=
\{\p:\Do^2\rightarrow \Re^+ | \forall \s\in\Do^2, \p(\s) \geq 0; \int_{\s\in\Do^2} \p(\s) d\s=1\}  \ .
\end{align*}
The space of $\p$ forms a Riemannian manifold, also known as the statistical manifold, which is well-known from the field of {\it information geometry} \citep{Amari85}. \citet{Rao:BCMS45} introduced the notion of the statistical manifold whose elements are probability density functions and composed the Riemannian structure with the {\it Fisher-Rao} metric. \citet{Cencov82} showed that the Fisher-Rao metric is the {\it unique intrinsic metric} on the statistical manifold $\P$ and therefore invariant to re-parameterizations of the functions. There are many different parameterizations of PDFs that are equivalent but with different forms of the Fisher-Rao metric, leading to the Riemannian operations with different computational complexity. In our study, we choose the square-root representation, which was used recently in ODF processing and registration \citep{Du:TMI_2011,Goh:NeuroImage2011,Cheng:MICCAI09}. The square-root representation is one of the most efficient representations found to date as the various Riemannian operations, such as geodesics, exponential maps, and logarithm maps, are available in closed form. 

The {\it square-root ODF} ($\gsODF$) is defined as
$\displaystyle
\bpsi(\s)=\sqrt{\p(\s)}$, where $\bpsi(\s)$ is assumed to be non-negative to ensure uniqueness. The space of such functions is defined as
\begin{align*}
\bPsi=
\{\bpsi:\Do^2\rightarrow \Re^+ | \forall \s\in\Do^2, \bpsi(\s) \geq 0; \int_{\s\in\Do^2} \bpsi^2(\s) d\s=1\}.
\end{align*}
We see that the functions $\bpsi$ lie on the positive orthant of a unit Hilbert sphere, a well-studied Riemannian manifold. It can be shown \citep{Srivastava:CVPR07} that the Fisher-Rao metric is simply the $\bL^2$ metric, given as
\begin{align*}
\langle \bxi_j, \bxi_k \rangle_{\bpsi_i} = 
\int_{\s\in\Do^2} \bxi_j(\s) \bxi_k (\s) d\s,
\end{align*}
where $\bxi_j, \bxi_k \in T_{\bpsi_i} \bPsi$ are tangent vectors at $\bpsi_i$. The geodesic distance between  any two functions $\bpsi_i, \bpsi_j \in \bPsi$ on a unit Hilbert sphere is the angle
\begin{align}
\label{eq:ODFdist}
\dist(\bpsi_i,\bpsi_j)=
\|\log_{\bpsi_i}(\bpsi_j)\|_{\bpsi_i} 
=\cos^{-1} \langle \bpsi_i, \bpsi_j \rangle = 
\cos^{-1}\left(\int_{\s\in\Do^2} \bpsi_i(\s) \bpsi_j(\s)d\s\right),
\end{align}
where $\langle \cdot, \cdot \rangle$ is the normal dot product between points in the sphere under the $\bL^2$ metric. For the sphere, the {\it exponential map} has the closed-form formula
\begin{align*}
\exp_{\bpsi_i}(\bxi) = \cos (\|\bxi\|_{\bpsi_i}) \bpsi_i + \sin(\|\bxi\|_{\bpsi_i})\frac{\bxi}{\|\bxi\|_{\bpsi_i}},
\end{align*}
where $\bxi \in T_{\bpsi_i}\bPsi$ is a tangent vector at $\bpsi_i$ and 
$\|\bxi\|_{\bpsi_i}=\sqrt{\langle \bxi,\bxi\rangle_{\bpsi_i}}$. By restricting $\|\bxi\|_{\bpsi_i}\in [0,\frac{\pi}{2}]$, we ensure that the exponential map is bijective. The {\it logarithm map} from $\bpsi_i$ to $\bpsi_j$ has the closed-form formula
\begin{align*}
\overrightarrow{\bpsi_i \bpsi_j}&=\log_{\bpsi_i}(\bpsi_j)
=\frac{\bpsi_j - \langle \bpsi_i,\bpsi_j \rangle \bpsi_i}{\sqrt{1-\langle \bpsi_i,\bpsi_j\rangle^2}} \cos^{-1} \langle \bpsi_i,\bpsi_j \rangle.
\end{align*}

\section*{References}
{\small
\bibliographystyle{elsarticle-harv}
\addcontentsline{toc}{section}{References}
\bibliography{ieeetmi_hardireg}
\label{journalend}
}

\end{document}

%% file: HARDI_NIMG2011_macro.tex
\newcommand{\dist}{\text{dist}}
\newcommand{\bPsi}{\bm{\Psi}}

\newcommand{\bOmega}{\bm{\Omega}}

\newcommand{\bpsi}{\bm{\psi}}

\newcommand{\bxi}{\bm{\xi}}

\renewcommand{\P}{\bm{\mathcal{P}}}

\newcommand{\Do}{\mathbb{S}}
\newcommand{\p}{\bm{p}}

\renewcommand{\Re}{{\mathbb{R}}}
\newcommand{\ie}{{i.e., }}

\newtheorem{property}{Property}
\newtheorem{assumption}{Assumption}

\DeclareMathOperator*{\argmax}{arg\,max}
\DeclareMathOperator*{\argmin}{arg\,min}

\newcommand{\Real}{\mathbb{R}}

\newcommand{\Id}{{\mathtt{Id}}}

\newcommand{\bL}{\mathbb{L}}
\newcommand{\G}{\mathcal{G}}

\newcommand{\s}{{\bf s}}

\newcommand{\bg}{\mathbf{g}}

\newcommand{\gsODF}{\sqrt{\text{ODF}}}

\newcommand{\atlas}{\text{atlas}}
\newcommand{\old}{\text{old}}
\newcommand{\new}{\text{new}}

\floatstyle{ruled}
\newfloat{alg}{thp}{lop}
\floatname{alg}{Algorithm}
\renewenvironment{algorithm}
{\begin{alg} \vspace{0cm} \noindent } {\vspace{0cm}
\end{alg}}